\documentclass[twoside,twocolumn]{article}
\usepackage{graphicx}
\usepackage{blindtext} 

\usepackage[sc]{mathpazo} 
\usepackage[T1]{fontenc} 
\usepackage{microtype} 

\usepackage[english]{babel} 

\usepackage[hmarginratio=1:1,top=32mm,columnsep=20pt]{geometry} 
\usepackage[hang, small,labelfont=bf,up,textfont=it,up]{caption} 
\usepackage{booktabs} 

\usepackage{lettrine} 

\usepackage{enumitem} 
\setlist[itemize]{noitemsep} 

\usepackage{abstract} 

\usepackage{titlesec} 
\renewcommand\thesection{\Roman{section}} 
\renewcommand\thesubsection{\roman{subsection}} 
\titleformat{\section}[block]{\large\scshape\centering}{\thesection.}{1em}{} 
\titleformat{\subsection}[block]{\large}{\thesubsection.}{1em}{} 

\usepackage{fancyhdr} 
\usepackage{titling} 

\usepackage{hyperref} 
\usepackage[noabbrev]{cleveref}
\usepackage[square,sort,comma,numbers]{natbib}
\usepackage[title]{appendix}

\pretitle{\begin{center}\Huge\bfseries} 
\posttitle{\end{center}} 
\title{The Atari Data Scraper} 
\author{%
\textsc{Brittany Davis Pierson}\thanks{Corresponding author} \\[1ex] 
\normalsize Washington State University \\ 
\normalsize \href{mailto:brittany.f.davis@wsu.edu}{brittany.f.davis@wsu.edu} 
\and 
\textsc{Justine Ventura} \\[1ex] 
\normalsize University of Alberta \\ 
\normalsize \href{mailto:linnrose@ualberta.ca}{linnrose@ualberta.ca} 
\and 
\textsc{Matthew E. Taylor} \\[1ex] 
\normalsize University of Alberta \\ 
\normalsize \href{mailto:matthew.e.taylor@ualberta.ca}{matthew.e.taylor@ualberta.ca} 
}
\date{} 
\renewcommand{\maketitlehookd}{%
\begin{abstract}
\noindent Reinforcement learning has made great strides in recent years due to the success of methods using deep neural networks. However, such neural networks act as a black box, obscuring the inner workings. While reinforcement learning has the potential to solve unique problems, a lack of trust and understanding of reinforcement learning algorithms could prevent their widespread adoption. Here, we present a library that attaches a ``data scraper'' to deep reinforcement learning agents, acting as an observer, and then show how the data collected by the Atari Data Scraper can be used to understand and interpret deep reinforcement learning agents. The code for the Atari Data Scraper can be found here: \url{https://github.com/IRLL/Atari-Data-Scraper}.
\end{abstract}
}

\begin{document}

\maketitle


\section{Introduction}

\lettrine[nindent=0em,lines=3]{R} einforcement learning allows an agent to learn from interacting with an environment iteratively, learning sequences of actions in order to perform tasks and reach goals \citep{suttonReinforcementLearningIntroduction2018a}. As reinforcement learning algorithms have achieved new records on benchmarks, they have also become more complex; many of the top-performing reinforcement learning algorithms today use deep neural networks, which are considered black-box algorithms.  It is unlikely that significantly less complex or more transparent reinforcement learning algorithms will be able to achieve the same performance, in part because reinforcement learning incorporates the temporal aspects of problem solving. Additionally, reinforcement learning algorithms perform best when they have access to more information about the environment. Consequently, top-performing algorithms to have the capability to represent complex features, often in non-linear ways. 

As deep reinforcement learning algorithms approach and surpass human capabilities in some domains, a new approach to interpreting and explaining such agents may be needed. Suppose we treat deep reinforcement learning algorithms as if they were human-like subjects. In doing so, we could use select methodologies from studies of human subjects in sociology, marketing, psychology, environmental sciences and more. In biological field research, for example, animals under observation may be fitted with a device to record specific information. As fitness trackers have developed, some sociological studies have similarly asked participants to wear devices to track information like daily activity levels. 

Current commonly-used reinforcement learning libraries like OpenAI's Baselines\footnote{\url{https://github.com/openai/baselines}}, Tensorforce\footnote{\url{https://github.com/openai/baselines} }, keras-rl\footnote{\url{https://github.com/keras-rl/keras-rl}}, TF-Agents\footnote{\url{https://www.tensorflow.org/agents}} and more make it difficult and time-consuming to collect additional data about deep reinforcement learning agents beyond simply their scores. However, we argue, improving access to data regarding the agent's actions, rewards, location, etc. could help to make deep reinforcement learning agents more interpretable. Consequently, users from novices to experts get more out of each time an agent interacts with an environment. In order to facilitate a movement towards more data-availability, we have created a function that creates an Atari Data Scraper. The Atari Data Scraper collects information about agents as they interact with various games from the Atari 2600 suite of OpenAI's gym \citep{brockmanOpenAIGym2016b} and saves the data for later examination. 

The remainder of this article is structured as follows: Section II covers relevant background on interpretable deep reinforcement learning. Section III details the creation of a data scraper for use with the popular Stable Baselines deep reinforcement learning library. Section IV features the outcome from running two agents with an Atari Data Scraper attached. Section V concludes with a discussion of the Atari Data Scraper's known limitations and the resulting directions for future work.


\section{Background}

Deep reinforcement learning's use of deep neural networks as function approximators makes it inherently difficult to interpret. The first algorithm to successfully learn from pixel-only input was the Deep Q-Network algorithm proposed by Mnih et al. in 2015 \citep{mnihHumanlevelControlDeep2015}. The algorithm, which is today referred to as DQN, consists of the original DQN algorithm with three improvements, all of which were proposed within a year of the groundbreaking paper's publication. In the paper advocating for one of those three improvements, Wang et al. used saliency maps to visualize how Dueling Q-Networks altered the functioning of the original DQN algorithm \citep{wangDuelingNetworkArchitectures2016}, showing that from the very start of deep reinforcement learning, visualizations were being used to interpret algorithms and their resulting agents. Saliency maps have played a large role in this, with some researchers examining in detail how well various existing saliency-creation methods work in deep reinforcement learning \citep{huberBenchmarkingPerturbationbasedSaliency2021} \citep{rosynskiAreGradientbasedSaliency2020}. Others are proposing entirely new methods tailored specifically to the challenges and capabilities of deep reinforcement learning algorithms \citep{greydanusVisualizingUnderstandingAtari2018a} \citep{rosynskiAreGradientbasedSaliency2020} \citep{nikulinFreeLunchSaliencyAttention2019} \citep{puriExplainYourMove2020}. 

However, visual inspection of saliency maps is not the only way deep reinforcement learning algorithms have been interpreted. There has long been a desire for more quantitative ways to interpret deep reinforcement learning agents. In a 2013 preprint of a less-developed DQN algorithm, Mnih et al. included a discussion about whether averaged reward or the estimated action-value function should be graphed to examine agent improvements and compare different agents \citep{mnihPlayingAtariDeep2013}. Since then, in many published landmark deep reinforcement learning papers, graphs are prominently featured as figures designed to provide a summary of the claims in the papers. Even saliency maps have been compared using distance measurements and other quantitative evaluation metrics rather than pure visual inspection \citep{bylinskiiWhatDifferentEvaluation2017}. 

As deep reinforcement learning has developed, the search for quantitative metrics has similarly evolved. In 2020, Sequeira et al. proposed a framework using records from an agent's interaction with an environment to make a more interpretable agent \citep{sequeiraInterestingnessElementsExplainable2020}. They showed how the framework might be used on a traditional Q-learning agent, but plan to extend the framework so that it can be used for DRL agents, too. The framework collected data such as how often a state was visited, current estimates of the values of states and actions, and the agent's uncertainty in each state. By collecting such data, the authors were able to do things like finding sequences from bad situations to good situations, or locate places in the environment where the agent is generally uncertain. 


In a similar vein, our Atari Data Scraper project seeks to uncover what could be learned by collecting data on a deep reinforcement learning agent, just as one might collect data on an animal species of interest in the wild. As a proof-of-concept, the first iteration of a data scraper was created as an additional class inserted into a fork of the OpenAI Baselines repository \citep{dhariwalOpenAIBaselines2017}. We modified the implementation of the Deep Q-Network (DQN) algorithm within the original repository to use this additional class to collect a set of data. After each time step, the data scraper object was given the following: 
\begin{itemize}
  \item The number associated with the action taken at that time step
  \item The reward received from the environment
  \item An array of the sum total of each action for the current life
  \item An array of the sum total of each action for the current game
  \item A boolean indicating if a life had been lost on this step
  \item The number of lives remaining
  \item An array of the $Q$ values for this step
  \item The feature vector describing the current game describing the current game screen 
\end{itemize}

The first iteration of the data scraper concept used this dataset to create a record for the current step. For all data passed in, besides the current $Q$ values and the feature vector, the data was added as-is to the step record. Additional data was also collected to aid in creating visualizations. These include a flag to denote if the step was the end of a game, the running total reward for the current life and the current game, and the running total of steps taken for the current life and the current game
Additionally, the feature vector was used to extract the x- and y-coordinates for Ms. Pacman and each of the four ghosts, which we will refer to as the characters in the environment. Then this data was used to calculate the distance between Ms. Pacman and each ghost, and the distance between Ms. Pacman and each of the four Power Pills. This data was in turn used to determine if Ms. Pacman had just eaten a Power Pill and, if so, mark that Power Pill as consumed for this game. 

\section{Implementation Details}

\subsection{The Initial Concept}

In the initial implementation, the characters' locations were found by first applying an edge detector to each game frame and then searching each discovered contour for a set of colors associated with each of five characters: Ms. Pacman and each of the four ghosts \citep{milliganDetectingObjectsPacMan2015}. If a color in the specified set was found, the center of that contour would be calculated and recorded as that character's position. The set of colors was created using an eyedropper tool on screenshots of the environment to determine as small a range of colors as possible for each character. 

\subsection{A More Generic Atari Data Scraper}

The initial proof-of-concept had some limitations. The Atari Data Scraper is a redesign of the initial proof-of-concept, and is available here: \url{https://github.com/IRLL/Atari-Data-Scraper}. In the initial implementation, a lot of computation was done on each step to create a complete record before the next step was taken.  In the second iteration of the Atari Data Scraper, this initial data collection was slimmed down to only the bare essentials to reduce computational ``drag.'' Then, the Atari Data Scraper can call a secondary program that takes that data and expands it from the saved file to a larger dataset. Alternatively, that processing can be attached to the end of an agent's run by passing in an optional flag.

Part of reducing the work done by the Atari Data Scraper during the agent's training was making the choice to save a screenshot of the current step rather than use the observed feature vector to immediately calculate the positions of the characters in the environment. While this choice greatly increased the storage used by the Atari Data Scraper, it allowed the data to be processed in two steps if needed. In order to handle the large number of images created by multiple runs of the Atari Data Scraper, an option was added to have the images automatically deleted after processing is completed. In addition, some of the calculations done by the first iteration of the Atari Data Scraper were not useful in understanding the agent. These calculations can still be done manually using the collected data, but we chose to scale back the total amount of data automatically generated and calculated to only that which was helpful in understanding our agents. Items marked with an asterisk in \cref{tab:data} are collected at each step by the Atari Data Scraper, while all other items are calculated or pulled using image processing by the secondary program.

\begin{table*}[h]
\caption{A listing of the data collected for DQN, A2C and PPO2 in the Ms. Pacman and Pong environments. An asterisk (*) marks data collected by default. All other data is collected and created in an optional second pass.}
\resizebox{\textwidth}{!}{%
\begin{tabular}{llll}
\multicolumn{4}{c}{Data Collected by the Atari Data Scraper}                                                                                                                                           \\ \hline
\multicolumn{2}{|c|}{Ms. Pacman}                                                                     & \multicolumn{2}{c|}{Pong}                                                                       \\ \hline
\multicolumn{1}{|c|}{A2C/PPO2}                    & \multicolumn{1}{c|}{DQN}                         & \multicolumn{1}{c|}{A2C/PPO2}                  & \multicolumn{1}{c|}{DQN}                       \\ \hline
\multicolumn{1}{|l|}{step number*}                & \multicolumn{1}{l|}{step number*}                & \multicolumn{1}{l|}{step number*}              & \multicolumn{1}{l|}{step number*}              \\ \hline
\multicolumn{1}{|l|}{action name*}                & \multicolumn{1}{l|}{action name*}                & \multicolumn{1}{l|}{action name*}              & \multicolumn{1}{l|}{action name*}              \\ \hline
\multicolumn{1}{|l|}{action number*}              & \multicolumn{1}{l|}{action number*}              & \multicolumn{1}{l|}{action number*}            & \multicolumn{1}{l|}{action number*}            \\ \hline
\multicolumn{1}{|l|}{step reward*}                & \multicolumn{1}{l|}{life reward*}                & \multicolumn{1}{l|}{game reward*}              & \multicolumn{1}{l|}{game reward*}              \\ \hline
\multicolumn{1}{|l|}{lives*}                      & \multicolumn{1}{l|}{lives*}                      & \multicolumn{1}{l|}{ball x-coordinate}         & \multicolumn{1}{l|}{ball x-coordinate}         \\ \hline
\multicolumn{1}{|l|}{characters' x-coordinates}   & \multicolumn{1}{l|}{characters' x-coordinates}   & \multicolumn{1}{l|}{ball y-coordinate}         & \multicolumn{1}{l|}{ball y-coordinate}         \\ \hline
\multicolumn{1}{|l|}{characters'   y-coordinates} & \multicolumn{1}{l|}{characters'   y-coordinates} & \multicolumn{1}{l|}{paddles'   x-coordinates}  & \multicolumn{1}{l|}{paddles'   x-coordinates}  \\ \hline
\multicolumn{1}{|l|}{distances to   ghosts}       & \multicolumn{1}{l|}{distances to ghosts}         & \multicolumn{1}{l|}{paddles'   y-coordinates}  & \multicolumn{1}{l|}{paddles'   y-coordinates}  \\ \hline
\multicolumn{1}{|l|}{pill eaten   statuses}       & \multicolumn{1}{l|}{pill eaten statuses}         & \multicolumn{1}{l|}{paddle to ball   distance} & \multicolumn{1}{l|}{paddle to ball   distance} \\ \hline
\multicolumn{1}{|l|}{distance to   pills}         & \multicolumn{1}{l|}{distance to pills}           & \multicolumn{1}{l|}{step reward}               & \multicolumn{1}{l|}{}                          \\ \hline
\multicolumn{1}{|l|}{current life   rewards}      & \multicolumn{1}{l|}{step reward}                 & \multicolumn{1}{l|}{}                          & \multicolumn{1}{l|}{}                          \\ \hline
\multicolumn{1}{|l|}{current game   rewards}      & \multicolumn{1}{l|}{current life step}           & \multicolumn{1}{l|}{}                          & \multicolumn{1}{l|}{}                          \\ \hline
\multicolumn{1}{|l|}{current life   step}         & \multicolumn{1}{l|}{current game step}           & \multicolumn{1}{l|}{}                          & \multicolumn{1}{l|}{}                          \\ \hline
\multicolumn{1}{|l|}{current game   step}         & \multicolumn{1}{l|}{game number}                 & \multicolumn{1}{l|}{}                          & \multicolumn{1}{l|}{}                          \\ \hline
\multicolumn{1}{|l|}{game number}                 & \multicolumn{1}{l|}{total reward}                & \multicolumn{1}{l|}{}                          & \multicolumn{1}{l|}{}                          \\ \hline
\multicolumn{1}{|l|}{total reward}                & \multicolumn{1}{l|}{life number}                 & \multicolumn{1}{l|}{}                          & \multicolumn{1}{l|}{}                          \\ \hline
\multicolumn{1}{|l|}{life number}                 & \multicolumn{1}{l|}{reward at end of game}       & \multicolumn{1}{l|}{}                          & \multicolumn{1}{l|}{}                          \\ \hline
\multicolumn{1}{|l|}{end of game   flag}          & \multicolumn{1}{l|}{}                            & \multicolumn{1}{l|}{}                          & \multicolumn{1}{l|}{}                          \\ \hline
\end{tabular}}
\label{tab:data}
\end{table*} 

We also used the redesign as an opportunity to improve the capabilities of the character tracking component of the Atari Data Scraper. For one, due to the low resolution of the game screen, the contours found by the edge detector were not always crisp. Occasionally, one break in a detected edge would lead to the contour determined to be Ms. Pacman to merge with a nearby wall into a single contour. When this happened, the center of the contour, which was stored as the character's location, would suddenly jump away and then return to a nearby location on the next step. To overcome this problem in the more generic version of the Atari Data Scraper, the function which locates characters does not use contours. Instead, it searches the entire image for the specified color ranges. Since each of the five characters is a different color when the ghosts are not toggled to dark-blue by a Power Pill, all five characters could be distinguished most of the time. Since Ms. Pacman never changes color, the agent's location could almost always be tracked using this method. If the ghosts are not found, then the contours are searched to try and locate four dark-blue ghosts. Secondly, the use of a color picker was phased out, as color pickers showed values which, despite being reported for the same color, were different on different devices. Instead, the matplotlib library was used to identify the color of specific pixels, and the reported colors were used to define the color ranges for each character, resulting in more accurate character tracking. 

The first iteration worked by inserting a function call within the implementation of DQN, at each step. This approach requires copying the repository so that the function call can be called after the agent takes each step in the environment. In order to allow the Atari Data Scraper to be used without having to customize a repository, we decided to choose a popular code base of deep reinforcement learning implementations and then developed a method which can be passed in to that library's existing functions. The OpenAI Baselines repository has little documentation, and has seen a drop-off in activity in recent years. Therefore, we chose a popular fork of OpenAI baselines called Stable Baselines \citep{hillStableBaselines2018}, which today is often recommended above Baselines. 

One useful feature of Stable Baselines is the ability to create custom callbacks. According to the Stable Baselines documentation, ``A callback is a set of functions that will be called at given stages of the training procedure \cite{hillStableBaselinesDocs}.'' The custom callback class provided by Stable Baselines includes a function which is called on each step. Using this function, we collect the slimmed-down data on each step. A callback can be used with an existing implementation of a reinforcement learning algorithm by simply setting the callback parameter in the learning function. This allows the improved second iteration of the Atari Data Scraper to be easily used with existing Stable Baselines implementations.


\section{Results}

The initial iteration of the Atari Data Scraper only worked with the DQN implementation and the Ms.Pacman environment. By using the callback method, the improved Atari Data Scraper could easily be passed into any reinforcement learning algorithm's implementation in the Stable Baselines library. We have tested and verified that the callback works with the Stable Baselines implementations of DQN, Advantage Actor Critic (A2C), and Proximal Policy Optimization (PPO2). We adjusted the callback to ensure it would work, not only for the Ms. Pacman environment, but also for the Pong environment. In the Pong environment, the two paddles and the ball are treated as characters, and their locations recorded at each time step along with the score and game number. In the Ms. Pacman environment, Ms. Pacman and the four ghosts are treated as characters. For both the Ms.Pacman environment and the Pong environment, we used the fourth version available through OpenAI gym. For each of the two environments, all three algorithms were run: PPO2, A2C, and DQN.  A notebook with the visualizations made using the data collected by the Atari Data Scraper can be found in the Atari Data Scraper repository. 

\begin{figure*}[ht] 
   \begin{centering}
   \includegraphics[width=5.15in]{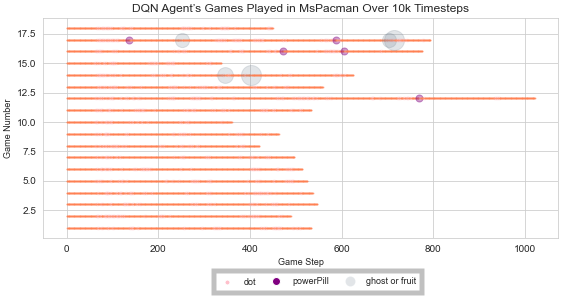} 
   \caption{A summary of an agent trained with DQN playing Ms. Pacman}
   \label{fig:dqn_training}
   \end{centering}
\end{figure*}

\begin{figure*}[ht] 
   \centering
   \includegraphics[width=5.8in]{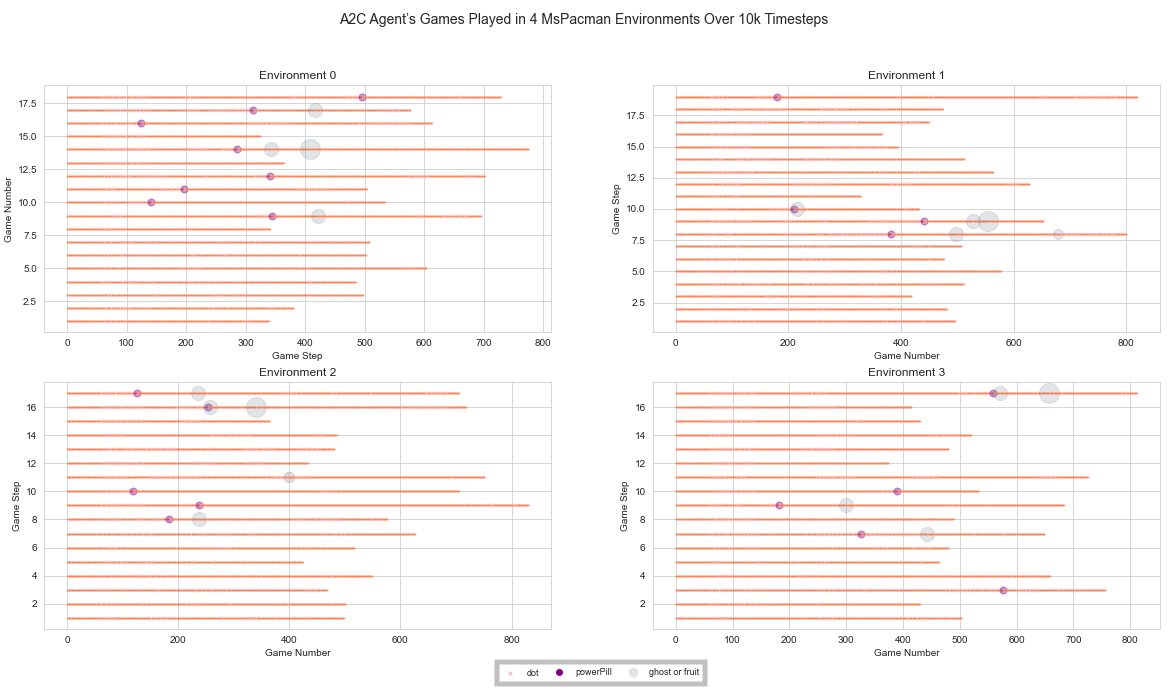} 
   \caption{A summary of an agent trained with A2C playing Ms. Pacman}
   \label{fig:4_env}
\end{figure*}

\begin{figure}[ht] 
   \centering
   \includegraphics[width=2.68in]{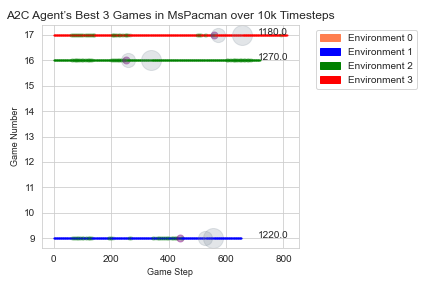} 
   \caption{The 3 bestgames played by an agent trained by the A2C algorithm using 4 environments}
   \label{fig:4_env_best}
\end{figure}

\begin{figure}[ht] 
   \centering
   \includegraphics[width=2.68in]{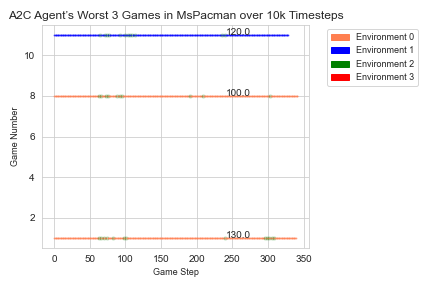} 
   \caption{The 3 worst games played by an agent trained by the A2C algorithm using 4 environments}
   \label{fig:4_env_worst}
\end{figure}

The DQN algorithm runs an agent through an environment, and data is collected about the characters in that environment. A2C and PPO2 can use multiple agents, each in its own environment, to do batch training. In situations with multiple environments running concurrently, the Atari Data Scraper collects information for each environment. For an agent trained to play the game Ms.Pacman using DQN, the information collected can be used to create a summary of all the games played by the agent, as shown in \cref{fig:dqn_training}. In addition, we used the Atari Data Scraper with agents trained to play the game Ms. Pacman with both the A2C and PPO2 algorithms. For the agent trained using the A2C algorithm, the collected data was used to examine the quality of the games played in each of 4 concurrent environments over time as seen in \cref{fig:4_env}. In \cref{fig:4_env_best} and \cref{fig:4_env_worst}, we show how the collected data can be filtered to display the top 3 best and worst performing games, with additional information about each game. 

\begin{figure}[ht] 
   \centering
   \includegraphics[width=2.8in]{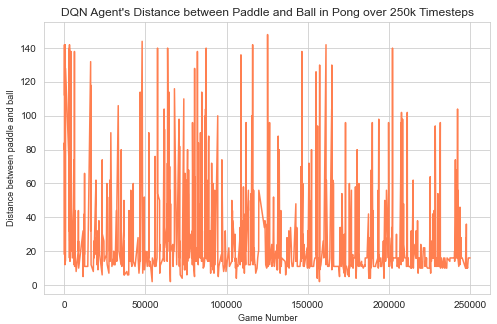} 
   \caption{The distance between the paddle and ball when the agent misses and the opponent gains a point. We can see it decreases on average as training increases.}
   \label{fig:pong_dist}
\end{figure}

\begin{figure}[ht] 
   \begin{centering}
   \includegraphics[width=2.8in]{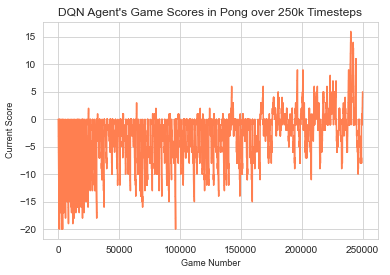} 
   \caption{The increasing ability of an agent trained with DQN to play Pong}
   \label{fig:pong_score}
   \end{centering}
\end{figure}

In Pong, the resulting data allowed us to examine the change in the distance between the agent's paddle and the ball each time the agent missed the ball, allowing the opponent to score a point, as seen in \cref{fig:pong_dist}. A game of Pong ends whenever the two players' scores differ by 20 points. Using the Atari Data Scraper, we were also able to visualize the agent's learning process in a more traditional manner, via the agent's score at the end of each game. The plot in \cref{fig:pong_score} starts consistently negative, as the opponent continues to get 20 more points than the agent quickly and easily, but gets into positive numbers more often as the agent learns how to score against the opponent. 

\section{Limitations and Future Work}

In an ideal world, we could find a data collection mechanism that could be used across all existing or forthcoming deep reinforcement learning algorithms. It would likewise be applicable to all environments, be it part of the OpenAI gym framework or a custom environment. In reality, any mechanism of collecting data from a deep reinforcement learning agent needs an attachment into either the agent or the environment by which it can access and collect the data. Thus, we had to impose limitations on the Atari Data Scraper. Any one of these imposed limitations is a potential avenue for future work. 

For one, there are some existing algorithm implementations which learn faster and have additional bells and whistles. In this project, we wanted to create a data collection mechanism that could be used easily by someone who is just trying to get their first deep reinforcement learning agent running, and also by a researcher who has worked in the field of deep reinforcement learning for years. For this reason, we avoided single-algorithm deep reinforcement learning implementations. Instead, we chose from among deep reinforcement learning libraries which have many of the most common algorithms implemented, any of which can be trained with a generic function call. The Stable Baselines library was chosen in part because it has more documentation and fewer bugs than other options we explored, creating a shallower learning curve for those just getting started in deep reinforcement learning. The actual collection process is set in motion by a function call specific to the Stable Baselines library, the callback. A clear limitation of the existing Atari Data Scraper is that  is that this specific function call would not work in other deep reinforcement libraries.

The Atari Data Scraper was also specifically designed around Atari environments within the OpenAI gym framework. We chose to focus on this set of environments because it was less limiting: the popular gym framework generalizes much of deep reinforcement learning, and so an Atari-focused Data Scraper could more easily be generalized to a multitude of other existing environments which are registered in the OpenAI gym. In comparison, a data scraper designed around a single custom environment, especially one not built around the gym framework, would need significant re-writing to be adapted to any other environment. To that end, the data collection performed by the Atari Data Scraper could also be achieved by an OpenAI gym wrapper. Such wrappers also allow access to a function called on each step. Future work which implements the Atari Data Scraper as a gym wrapper would allow the Data Scraper to be generalized across deep reinforcement learning libraries and potentially even to some of the most cutting-edge algorithms. A data collection mechanism written as a wrapper would still be limited to only environments registered in the OpenAI gym. Other future work could choose instead to write data collection mechanisms for some of the most popular environments for deep reinforcement learning outside of the gym framework. 

However, this would still leave one of the most restrictive limitations on the current Atari Data Scraper in place. The biggest limitation is the method by which characters and items in the environment are located. In order to locate characters, the Atari Data Scraper saves a screenshot of each step, requiring a lot of storage space. The screenshot is then searched, using image processing to find pixels within hand-crafted color ranges. Thus, to be applied to a new environment, the Atari Data Scraper must be equipped with painstakingly calculated color ranges for individual objects in the environment. Logically, this also limits the Atari Data Scraper to environments where the color of important objects does not appear elsewhere in the environment. Future work could include developing a method of automatically tracking items in the environment, ideally in a way that also allows broader generalization across the gym framework and across various libraries and implementations. 

Finally, the Atari Data Scraper uses the reward signal to infer other information. For example, in Ms. Pacman, eating a Power Pill earns the agent a reward of positive fifty. The Atari Data Scraper uses this fact, along with the agent's coordinates, to record when each Power Pill is eaten in a game. Thus, the Atari Data Scraper will not work as well for environments with clipped rewards or with rewards that do not differentiate between events. For example, in Ms. Pacman, eating one type of fruit and one type of ghost result in the same reward, making it much more difficult to determine which event occurred just by looking at the current reward signal. Additional future work could seek other ways of accessing the information currently carried in rewards signals. Alternatively, it could focus on transforming any rewards signal so that all important events provide differing rewards. 

\section{Acknowledgements}

Part of this work has taken place in the \href{https://irll.ca/}{Intelligent Robot Learning Laboratory at the University of Alberta (IRLL)}, supported in part by research grants from the Alberta Machine Intelligence Institute (Amii), CIFAR, and NSERC. We would like to acknowledge the help provided in this project by the members of  \href{https://irll.ca/}{IRLL}. We are honored to be researching alongside such hard-working people and grateful for all their help in developing these ideas, proof-reading our writing, and testing our code. We would especially like to thank Dr. Matthew Taylor for his assistance in putting this team together and his guidance throughout this project. In addition, we would like to thank Dr. Dustin Arendt at Pacific Northwest National Laboratory for his advice and guidance on this project.

\bibliographystyle{plain}
\bibliography{DataScraper_DavisVenturaTaylor} 

\begin{thebibliography}{10}

\bibitem{brockmanOpenAIGym2016b}
Greg Brockman, Vicki Cheung, Ludwig Pettersson, Jonas Schneider, John Schulman,
  Jie Tang, and Wojciech Zaremba.
\newblock {{OpenAI Gym}}.
\newblock {\em arXiv:1606.01540 [cs]}, June 2016.

\bibitem{bylinskiiWhatDifferentEvaluation2017}
Zoya Bylinskii, Tilke Judd, Aude Oliva, Antonio Torralba, and Fr{\'e}do Durand.
\newblock What do different evaluation metrics tell us about saliency models?
\newblock {\em arXiv:1604.03605 [cs]}, April 2017.

\bibitem{dhariwalOpenAIBaselines2017}
Prafulla Dhariwal, Christopher Hesse, Oleg Klimov, Alex Nichol, Matthias
  Plappert, Alec Radford, John Schulman, Szymon Sidor, Yuhuai Wu, and Peter
  Zhokhov.
\newblock {{OpenAI Baselines}}, 2017.

\bibitem{greydanusVisualizingUnderstandingAtari2018a}
Sam Greydanus, Anurag Koul, Jonathan Dodge, and Alan Fern.
\newblock Visualizing and {{Understanding Atari Agents}}.
\newblock {\em arXiv:1711.00138 [cs]}, September 2018.

\bibitem{hillStableBaselinesDocs}
Ashley Hill.
\newblock Stable {{Baselines Docs}}: {{Callbacks}}.

\bibitem{hillStableBaselines2018}
Ashley Hill, Christopher Hesse, Oleg Klimov, Alex Nichol, Matthias Plappert,
  Alec Radford, John Schulman, Szymon Sidor, Yuhuai Wu, Peter Zhokhov, Antonin
  Raffin, Maximilian Ernestus, Adam Gleave, Anssi Kanervisto, Rene Traore, and
  Prafrulla Dhariwal.
\newblock Stable {{Baselines}}, 2018.

\bibitem{huberBenchmarkingPerturbationbasedSaliency2021}
Tobias Huber, Benedikt Limmer, and Elisabeth Andr{\'e}.
\newblock Benchmarking {{Perturbation}}-based {{Saliency Maps}} for
  {{Explaining Deep Reinforcement Learning Agents}}.
\newblock {\em arXiv:2101.07312 [cs]}, January 2021.

\bibitem{milliganDetectingObjectsPacMan2015}
RD~Milligan.
\newblock Detecting objects in {{Pac}}-{{Man}} ({{Mark II}}), 2015.

\bibitem{mnihPlayingAtariDeep2013}
Volodymyr Mnih, Koray Kavukcuoglu, David Silver, Alex Graves, Ioannis
  Antonoglou, Daan Wierstra, and Martin Riedmiller.
\newblock Playing {{Atari}} with {{Deep Reinforcement Learning}}.
\newblock {\em arXiv:1312.5602 [cs]}, December 2013.

\bibitem{mnihHumanlevelControlDeep2015}
Volodymyr Mnih, Koray Kavukcuoglu, David Silver, Andrei~A. Rusu, Joel Veness,
  Marc~G. Bellemare, Alex Graves, Martin Riedmiller, Andreas~K. Fidjeland,
  Georg Ostrovski, Stig Petersen, Charles Beattie, Amir Sadik, Ioannis
  Antonoglou, Helen King, Dharshan Kumaran, Daan Wierstra, Shane Legg, and
  Demis Hassabis.
\newblock Human-level control through deep reinforcement learning.
\newblock {\em Nature}, 518(7540):529--533, February 2015.

\bibitem{nikulinFreeLunchSaliencyAttention2019}
Dmitry Nikulin, Anastasia Ianina, Vladimir Aliev, and Sergey Nikolenko.
\newblock Free-{{Lunch Saliency}} via {{Attention}} in {{Atari Agents}}.
\newblock {\em arXiv:1908.02511 [cs]}, October 2019.

\bibitem{puriExplainYourMove2020}
Nikaash Puri, Sukriti Verma, Piyush Gupta, Dhruv Kayastha, Shripad Deshmukh,
  Balaji Krishnamurthy, and Sameer Singh.
\newblock Explain {{Your Move}}: {{Understanding Agent Actions Using Specific}}
  and {{Relevant Feature Attribution}}.
\newblock {\em arXiv:1912.12191 [cs]}, April 2020.

\bibitem{rosynskiAreGradientbasedSaliency2020}
Matthias Rosynski, Frank Kirchner, and Matias {Valdenegro-Toro}.
\newblock Are {{Gradient}}-based {{Saliency Maps Useful}} in {{Deep
  Reinforcement Learning}}?
\newblock {\em arXiv:2012.01281 [cs]}, December 2020.

\bibitem{sequeiraInterestingnessElementsExplainable2020}
Pedro Sequeira and Melinda Gervasio.
\newblock Interestingness {{Elements}} for {{Explainable Reinforcement
  Learning}}: {{Understanding Agents}}' {{Capabilities}} and {{Limitations}}.
\newblock {\em Artificial Intelligence}, 288:103367, November 2020.

\bibitem{suttonReinforcementLearningIntroduction2018a}
Richard~S. Sutton and Andrew~G. Barto.
\newblock {\em Reinforcement Learning: An Introduction}.
\newblock Adaptive Computation and Machine Learning Series. {The MIT Press},
  {Cambridge, Massachusetts}, second edition edition, 2018.

\bibitem{wangDuelingNetworkArchitectures2016}
Ziyu Wang, Tom Schaul, Matteo Hessel, Hado {van Hasselt}, Marc Lanctot, and
  Nando {de Freitas}.
\newblock Dueling {{Network Architectures}} for {{Deep Reinforcement
  Learning}}.
\newblock {\em arXiv:1511.06581 [cs]}, April 2016.

\end{thebibliography}



\end{document}